\documentclass[final]{l4dc2026}
\usepackage{newunicodechar}
\newunicodechar{ℒ}{\mathcal{L}}

\title[$\mathcal{L}_1$-augmented dynamical systems]{A Robust Task-Level Control Architecture for Learned Dynamical Systems}
\usepackage{times}
\usepackage{times}
\usepackage{ulem}
\usepackage{mathtools,amsmath,amssymb,amsfonts}
\usepackage{bm}
\usepackage{enumitem}
\usepackage{hyperref}
\usepackage{xcolor}
\usepackage{float}
\usepackage{graphicx}
\usepackage{subcaption} 
\usepackage{float}
\usepackage{amsmath}
\usepackage{booktabs} 
\usepackage{multirow}
\usepackage{float}     
\usepackage{caption}   

\usepackage{hyperref}

\newcommand{\unom}{u_{\text{nom}}}

\newcommand{\mytextsize}{\fontsize{9pt}{9pt}\selectfont}

\newcounter{assumption}
\renewcommand{\theassumption}{A\arabic{assumption}}

\makeatletter
\newenvironment{assumption}[1][]%
{%
  \refstepcounter{assumption}%
  \par\smallskip
  \noindent\textbf{Assumption \theassumption%
  \if\relax\detokenize{#1}\relax\else\ (\textnormal{#1})\fi.}\ %
  \itshape
}%
{%
  \par\smallskip\normalfont
}
\makeatother

\coltauthor{\Name{Eshika Pathak$^\dagger$} \Email{epathak2@illinois.edu}\\
 \Name{Ahmed Aboudonia$^\ast$} \Email{aboudonia@berkeley.edu}\\
 \Name{Sandeep Banik$^\dagger$} \Email{baniksan@illinois.edu}\\
 \Name{Naira Hovakimyan$^\dagger$} \Email{nhovakim@illinois.edu}\\
 \addr $^\dagger$University of Illinois Urbana-Champaign, $^\ast$University of California, Berkeley}

\begin{document}

\maketitle

\begin{abstract}
Dynamical system (DS)-based learning from demonstration (LfD) is a powerful tool for generating motion plans in the operation (`task') space of robotic systems. However, realizing generated motion plans is often compromised by a ``task-execution mismatch'', where unmodeled dynamics, persistent disturbances, and system latency cause the robot's task-space state to diverge from the desired state. 
We propose a novel task-level robust control architecture, $\mathcal{L}_{1}$-augmented Dynamical Systems ($\mathcal{L}_{1}$-DS), that explicitly handles the task-execution mismatch in tracking a nominal motion plan generated by any DS-based LfD scheme. 
Our framework augments any DS-based LfD model with a nominal stabilizing controller and an $\mathcal{L}_{1}$ adaptive controller. 
Furthermore, we introduce a windowed Dynamic Time Warping (DTW)-based target selector, which enables the nominal stabilizing controller to handle temporal misalignment for improved phase-consistent tracking. We demonstrate the efficacy of our architecture on the LASA and IROS handwriting datasets.

\end{abstract}

\begin{keywords}%
  Learning from demonstrations; operation/task space; adaptive control; motion plans.
\end{keywords}


\section{Introduction}




The execution of complex
manipulation skills, such as part assembly in factories, and surface cleaning in homes, are essential for robots operating in the real world. However, programming these skills explicitly is challenging due to their inherent complexity. 
Learning from Demonstration (LfD) \citep{billard2008robot} has emerged as an effective method, enabling robots to acquire skills from expert demonstrations (e.g., human-guided kinesthetic teaching or teleoperation).
One such data-efficient approach within LfD encodes skills as autonomous dynamical systems (DS) \citep{fu2026survey}, and learns a function that maps the robot state to a desired velocity (by learning a vector field) in the operational (task) space \citep{khatib2003unified}.
The learned DS is then used to generate task-space motion plans. 
However, during execution, sensing delays, latency, unmodeled low-level dynamics, and persistent external disturbances, often create a ``task-execution gap'', where the robot's true motion diverges from the generated motion plan. In this work, we address the task-execution gap for trajectory-following tasks by designing stable and robust DS-based motion plans.




In DS-based LfD, various efforts have been devoted to ensuring the stability of the generated DS-based motion plans. 
Early methods, such as dynamic movement primitives \citep{pastor2009learning, saveriano2023dynamic}, model skills as stable second-order systems with learned time-dependent forcing terms. Time-invariant approaches, such as SEDS \citep{khansari2011learning}, learn globally asymptotically stable DS to a target point (`attractor') in task-space, but their strict stability constraints often reduce motion reproduction accuracy, as discussed in~\citep{saveriano2020energy, figueroa2018physically}. Related work has pursued both expressiveness through deep learning formulations: polynomial DS \citep{abyaneh2023learning}, Neural ODEs \citep{sochopoulos2024learning, nawaz2024learning},
and normalizing flows \citep{urain2020imitationflow}; and stability guarantees through methods such as data-driven Lyapunov candidates \citep{neumann2013neural, khansari2014learning}, contraction theory \citep{blocher2017learning}, and diffeomorphic transformations \citep{neumann2015learning, rana2020euclideanizing}. Further extensions include non-Euclidean geometries \citep{zhang2022learning, duong2021hamiltonian}, rhythmic motions \citep{sandor2015sensorimotor, nawaz2024learning} and multi-step tasks \citep{li2023task}.

The above methods provide formal stability guarantees for DS-based motion plans (via Lyapunov theory or contraction), and assume a ``perfect executor''- a low-level controller that tracks desired task-level motion plan perfectly, even in the presence of disturbances. 
To account for disturbances, many robust low-level controllers (using methods like adaptive robust feedback linearization \citep{abdelwahab2024adaptive}, backstepping super-twisting \citep{kali2021backstepping}, and adaptive coordination schemes 
\citep{khadivar2023adaptive}) can, in principle, be developed.  Their design, however, typically relies on the availability of a nominal system model, which can be difficult to obtain or utilize as modern robotic systems become increasingly complex \citep{khatib2004whole, brantner2025humanoid, lovett2025review, taha2012flight, he2020modeling}. Furthermore, most commercial robots are ``black boxes'' that restrict access to low-level control due to certifiability or intellectual property concerns \citep{bilancia2023overview}. 
Our proposed framework enables robust execution without requiring access to low-level control or precise system models.

Recent work in DS-based LfD has sought to augment or embed robustness directly into the learned dynamics.
For example, instantaneous state perturbations (e.g. a robot arm being suddenly dragged and dropped off-course to a new position in the task space) can be corrected online via a quadratic program to ensure stable and safe tracking of motion plans \citep{nawaz2024learning} . 
Other approaches include generating an online modulation policy for sequential single-attractor sub-tasks \citep{wang2024novel}, and achieving safety by learning control barrier functions \citep{saveriano2019learning, huang2025safety}. LAGS-DS \citep{figueroa2022locally}, on the other hand, learns a DS with global convergence and stiffness-like symmetric attraction around reference trajectories in task-critical regions, but requires manual specification of these regions and is restricted to single-global-attractor tasks. To track a desired motion plan, leveraging the dynamical system nature of the task-level planner, we employ control-theoretic tools and address the “task-execution mismatch”.

In this work, we propose $\mathcal{L}_1$-DS, an architecture that augments any learned DS-based LfD model with a nominal stabilizing controller, based on Control Lyapunov Functions (CLFs), and an $\mathcal{L}_1$ adaptive controller to actively handle the task-execution mismatch. Our approach is agnostic to the robot's low-level control stack, addressing the mismatch purely at the task-level. To improve the ability of the nominal stabilizing controller to handle temporal misalignments, we introduce a windowed Dynamic Time Warping (DTW)-based target selector.
We empirically validate this framework on the LASA and IROS handwriting datasets, demonstrating the efficacy of $\mathcal{L}_1$-DS 
under various types of disturbances.

\section{Background}
\label{sec:background}


\subsection{Learning Dynamical Systems from Demonstrations}
\label{sec:bg-ds}



A continuous-time DS-based motion plan for a robotic system is defined as a function of an unambiguous operational/task-space state $z_{\text{des}}(t)\in\mathbb{R}^d$ at time $t \in \mathbb{R}_{\geq0}$, which encodes the task-relevant quantities (e.g., end-effector pose in operation space \citet{khatib1986motion, mistry2012operational, lee2021whole}). The motion plan 
is characterized as an autonomous nonlinear system:
    $\dot{z}_{\text{des}}(t) \;=\; f(z_{\text{des}}(t)), 
    {\text{with}} \ z_{\text{des}}(t_0)=z_0,$
where $f:\mathbb{R}^d\to\mathbb{R}^d$ is an unknown smooth (time-invariant) vector field that encodes the desired specific motion primitives to be reproduced during execution from an initial state $x_{0}$ at time $t_0 \leq t$.
Given a dataset consisting of state--velocity pairs from $N$ demonstrations, $\mathcal{D} := \{z_i(t_k), \dot{z}_i(t_k)\}_{i,k}$ with $i \in \{1, \dots, N\}$, and $k \in \{1, \dots, T_i\}$ ($T_i$ being the length of the $i$-th demonstration), we learn the parameterized nominal dynamics $f_\theta : \mathbb{R}^d \to \mathbb{R}^d$. To find the optimal parameters $\theta^*$, the estimator $f_\theta$ is fit by minimizing a loss over $\mathcal{D}$,
$$
\theta^* \in \arg \min_\theta \sum_{i=1}^N \sum_{k=1}^{T_i} \mathcal{L}(z_i(t_k), \bar{z}_i(t_k), \dot{z}_i(t_k), f_\theta(z_i(t_k))), \text{ subject to } C(\theta) \le 0,
$$
where $\mathcal{L}:\mathbb{R}^d \times \mathbb{R}^d \times \mathbb{R}^d \times \mathbb{R}^d \to \mathbb{R}$ is a chosen loss function, and $\bar{z}_i(t_k)$ denotes the state obtained by integrating $\dot{\bar{z}} = f_\theta(\bar{z})$ from $t_0$ to $t_k$ given $\bar{z}(t_0) = z_i(t_0)$. 
The optional term $\mathcal{C}(\theta) \le 0$ represents inequality constraints imposed on the parameters to guarantee desired system properties (such as Lyapunov stability).
Methods of modeling and fitting $f_\theta$ include, but are not limited to, \citet{khansari2011learning,figueroa2018physically, sochopoulos2024learning, nawaz2024learning, figueroa2022locally, urain2020imitationflow}, and \citet{saveriano2020energy}.

\subsection{\texorpdfstring{$\mathcal{L}_1$}{L1} Adaptive Control}
\label{sec:bg-l1}

$\mathcal{L}_1$ adaptive control presents an architecture that decouples estimation from control and ensures uniform transient guarantees along with  apriori computable robustness margins for a broad class of uncertain systems~\citep{hovakimyan2010ℒ1}. Its effectiveness has been successfully demonstrated in different safety-critical deployments, including NASA's AirStar 5.5\% model \citep{gregory2010flight}, Calspan's Learjet \citep{ackerman2017evaluation}, unmanned aerial vehicles \citep{kaminer2010path, wu2023mathcal}  and robotic systems \citep{pravitra2020ℒ, cheng2022improving, sung2024robust}. We briefly review the $\mathcal{L}_1$ adaptive control architecture below. More details can be found in \citet{hovakimyan2010ℒ1}.

Consider an uncertain control-affine nonlinear system:
\begin{equation}
\label{eq:control-affine-nonlinear-bg}
\dot{x}(t) = g(x(t)) + h(x(t)) u(t) + \delta(t, x(t), u(t)), \quad x(t_0) = x_0,
\end{equation}
where $x(t) \in \mathbb{R}^d$ is the system state, $u(t) \in \mathbb{R}^m$ is the control input, $g: \mathbb{R}^d \to \mathbb{R}^d$ and $h: \mathbb{R}^d \to \mathbb{R}^{d \times m}$ are known functions, and $\delta: \mathbb{R}_{\geq 0} \times \mathbb{R}^d \times \mathbb{R}^m \to \mathbb{R}^d$ represents the unknown residual uncertainty. The uncertainty is decomposed into two components, namely, matched $\delta_m \in \mathbb{R}^m$ (acting through the control channel $h(x)$) and unmatched $\delta_{um} \in \mathbb{R}^{d-m}$ (acting through its orthogonal complement $h^\perp(x)$), i.e., $\delta(t, x, u) = h(x) \delta_m(t, x, u) + h^\perp(x) \delta_{um}(t, x, u).$
The control objective is to ensure the system state $x(t)$ remains uniformly bounded around a desired reference trajectory $x_{\text{ref}}(t)$, generated by the reference system $\dot{x}_{\text{ref}}(t) = g(x_{\text{ref}}(t)) + h(x_{\text{ref}}(t)) u_{\text{nom}}(t) \ \text{with} \ x_{\text{ref}}(t_0) = x_0$, where $u_{\text{nom}}(t)$ is generated by a nominal controller. 

In an ideal, uncertainty-free case ($\delta = 0$), applying $u(t) = u_{\text{nom}}(t)$ to the system yields perfect tracking. In the presence of uncertainty, the $\mathcal{L}_1$ adaptive controller adds an adaptive term $u_a(t)$ such that the total control input  $u(t) = u_{\text{nom}}(t) + u_a(t)$ achieves the aforementioned control objective.
The $\mathcal{L}_1$ adaptive control architecture consists of three main components:


\paragraph{a. State Predictor:}
The state predictor equation is given by:
\begin{equation}
\label{eq:state_predictor_bg}
\dot{\hat{x}}(t) = g(x(t)) + h(x(t)) u(t) + \hat{\delta}(t) + A_s \tilde{x}(t), \quad \hat{x}(t_0)= x_0,
\end{equation}
where $\tilde{x}(t) = \hat{x}(t) - x(t)$ is the prediction error, $A_s$ is a user-defined Hurwitz matrix, and $\hat{\delta}(t)$ is an estimate of the uncertainty calculated using the adaptation law stated below. A small prediction error $\tilde{x}(t)$ implies that $\hat{\delta}(t)$ accurately estimates the true uncertainty $\delta(t, x(t), u(t))$.

\paragraph{b. Piecewise-Constant Adaptive Law:}
The uncertainty estimate $\hat{\delta}(t)$ is updated at discrete sampling instants $t_i := i T_s$, $i \in \mathbb{N}$, with sampling period $T_s > 0$. The estimate is held constant between samples:
\begin{equation}
\label{eq:pc_adapt_law}
\hat{\delta}(t) = \hat{\delta}(t_i) = - \Phi^{-1}(T_s) \mu(t_i), \quad t \in [t_i, t_{i+1}),
\end{equation}
where
$\Phi(T_s) = A_s^{-1} ( e^{A_s T_s} - I ) \ \text{and} \ \mu(t_i) = e^{A_s T_s} \tilde{x}(t_i). $ 
This update law is designed to cancel the effect of prior prediction errors at the sampling instants, enabling fast and stable estimation.

\paragraph{c. Low-Pass Filtered Control Law:}
The pseudo-inverse of $h(x(t))$ is used to compute the matched component of the uncertainty $\hat{\delta}_m(t) = h(x(t))^+ \hat{\delta}(t),$
which is passed through a strictly proper, stable low-pass filter $C(s)$ to generate the adaptive control input:
\begin{equation}
u_a(s) = - C(s)\hat{\delta}_m(s),
\end{equation}
where $\hat{\delta}_m(s)$ is the Laplace transform of $\hat{\delta}_m(t)$, and $C(s) = \frac{\omega}{s + \omega}$ is the transfer function of a first-order low-pass filter with $\omega > 0$ as filter bandwidth chosen to satisfy the \(\mathcal{L}_1\) small-gain stability condition \citep{wang2011ℒ}. This filter is the key to decoupling adaptation from robustness, as it filters high-frequency estimation noise from entering the control channel and destabilizing the system.

\subsection{Control Lyapunov Function (CLF) Quadratic Programs}
\label{sec:bg-clf}



Consider the control-affine nonlinear system (\ref{eq:control-affine-nonlinear-bg}) in the absence of uncertainties (i.e., $\delta = 0$).
For $x \in \mathcal{X} \subseteq \mathbb{R}^{d}$, a continuously differentiable function \( V : \mathcal{X} \to \mathbb{R}_{\geq 0} \) is a CLF for $x=0$ if:
$
V(0) = 0, \ V(x) > 0 \ \text{for all } x \neq 0,
$
and there exists a constant \( c > 0 \) such that for all \( x \in \mathcal{X} \setminus \{0\} \),
$
\inf_{u \in \mathbb{R}^m} \left\{ L_g V(x) + L_h V(x) u + c V(x) \right\} \leq 0,
$
where
$
L_g V(x), L_h V(x) $
denote the Lie derivatives of \( V \) along \( g \) and \( h \), respectively \citep{khalil2015nonlinear}.
Given a candidate CLF, a control input \( \unom \) 
can be computed, at each time instant, as the solution to the quadratic program (QP):
$
\unom(x) = \arg\min_{u \in \mathbb{R}^m} \|u\|_2^2 \ , \text{s.t.} \ L_g V(x) + L_h V(x) u + c V(x) \leq 0$, to asymptotically stabilize the uncertainty-free system. 
This formulation can be extended using control barrier functions \citep{ames2019control} to enforce safety objectives.

\subsection{Dynamic Time Warping (DTW)}
\label{sec:bg-dtw}

DTW \citep{sakoe2003dynamic} is a technique for measuring the similarity between two temporal sequences that can vary in speed or timing. Given two sequences $A=(a_1,\ldots,a_M)$ and $B=(b_1,\ldots,b_N)$ of lengths $M$ and $N$, respectively, DTW finds an optimal alignment by constructing a warping path that minimizes the cumulative cost of local alignments. Let $c(i,j) := \|a_i - b_j\|_2$ denote the local distance between elements $a_i$ and $b_j$. The DTW algorithm employs dynamic programming to compute the cumulative distance matrix $D(i,j)$, defined by the recurrence relation:
\begin{align}
D(0,0) &= 0, \quad D(i,0) = D(0,j) = +\infty \quad \forall i,j > 0 \nonumber\\
D(i,j) &= c(i,j) + \min\{D(i-1,j), D(i,j-1), D(i-1,j-1)\}
\label{eq:bg-dtw-dp}
\end{align}
for $i \in \{1,\ldots,M\}$ and $j \in \{1,\ldots,N\}$. The DTW distance between sequences $A$ and $B$ is given by $\text{DTW}(A,B) = D(M,N)$, and the optimal warping path is recovered through backtracking from $(M,N)$ to $(1,1)$. The warping path $\mathcal{P} = \{(i_k, j_k)\}_{k=1}^K$ must satisfy three fundamental constraints: (i) \textit{boundary conditions} requiring $\mathcal{P}$ to begin at $(1,1)$ and end at $(M,N)$; (ii) \textit{monotonicity}, ensuring $i_{k+1} \geq i_k$ and $j_{k+1} \geq j_k$ for all $k$; and (iii) \textit{continuity}, restricting step transitions to $\{(1,0), (0,1), (1,1)\}$. To prevent pathological alignments and reduce computational complexity, the Sakoe-Chiba band constraint restricts the warping path to regions, where $|i-j| \leq w$ for some window parameter $w \geq 0$. This constraint reduces the time complexity from $O(MN)$ to $O(w \cdot \min\{M,N\})$, and avoids extreme warping distortions.




\section{Proposed Control Architecture: \texorpdfstring{$\mathcal{L}_1$}{L1}-DS}





\label{sec:method-arch}

\begin{figure}[t]
    \centering
    \includegraphics[width=0.7\columnwidth]{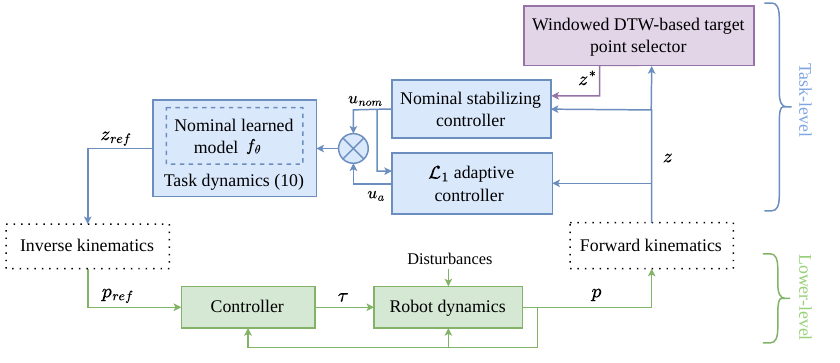}
    \caption{Proposed control architecture. $p_{\text{ref}}$ is the reference low-level state (e.g., desired joint positions), $\tau$ represents the control inputs (e.g., joint torques) applied to the robot dynamics, and $p$ is the actual measured low-level state (e.g., actual joint positions).}
    \label{fig:architecture}
\end{figure}

We now present our proposed control architecture, $\mathcal{L}_1$-DS, shown in Figure \ref{fig:architecture}, to generate a stable and robust DS-based motion plan.  We denote the robot's task-space state as $z(t) \in \mathbb{R}^d$. This state is expected to follow a nominal target trajectory $z^*(t)$, generated by the nominal learned dynamics $f_\theta:\mathbb{R}^d\to\mathbb{R}^d$ (see Section~\ref{sec:nominal-learned-dynamics}), which satisfies the initial value problem:
\begin{equation}
\label{eq:nom-target-traj}
    \dot{z}^*(t) = f_{\theta}(z^*(t)), \quad z^*(t_0) = z_0^\star.
\end{equation}
During execution, the actual state $z(t)$ will inevitably deviate from the target $z^*(t)$ due to the ``task-execution gap" caused by uncertainties such as external disturbances, latency, and modeling errors. 
For a given radius $\rho > 0$, we define a tube around $z^*(t)$ as $\mathcal{T}(\rho) \triangleq \bigcup_{t \ge t_0} \mathcal{O}(z^*(t), \rho)$, where $\mathcal{O}(z^*(t), \rho) := \{z \in \mathbb{R}^d \mid ||z - z^*(t)|| \le \rho\}$ (pointwise ball around $z^{*}(t)$).
The primary control objective is to guarantee that the actual state $z(t)$ remains bounded
around the nominal target trajectory $z^*(t)$, formally expressed as:
\begin{equation}
\label{eq:goal}
    z(t) \in \mathcal{O}(z^*(t), \rho), \quad \forall t \ge t_0.
\end{equation}
The $\mathcal{L}_1$-DS architecture consists of four components at the task-level: (i) the nominal learned dynamics $f_{\theta}$, defining the nominal motion plan, 
(ii) a nominal control block to assure the stability of the nominal learned dynamics, (iii) a windowed DTW-based target selector to maintain temporal alignment with respect to the nominal target trajectory, and (iv) an $\mathcal{L}_{1}$ adaptive control block that actively handles the task-execution mismatch.



\subsection{Nominal Learned Dynamics}
\label{sec:nominal-learned-dynamics}

The nominal dynamics, $f_{\theta}$, is learned from demonstrations (Section~\ref{sec:bg-ds}), and is rolled out to generate the nominal target trajectory $z^*(t)$ (\ref{eq:nom-target-traj}). 
To ensure that the control design and stability analysis are well-posed, we define the domain of operation and impose mild regularity assumptions on $f_{\theta}$.


\begin{assumption}[Operating domain and well-posedness]
\label{ass:domain}
There exists a compact, forward-invariant set $\mathcal{Z} \subset \mathbb{R}^d$ such that $T(\rho) \subseteq \mathcal{Z}$. 
The learned dynamics $f_\theta : \mathcal{Z} \to \mathbb{R}^d$ is locally Lipschitz continuous on $\mathcal{Z}$ and bounded: there exists $\Delta_f > 0$ such that
$
\|f_{\theta}(z)\| \le \Delta_f, \forall z \in \mathcal{Z}.
$
\end{assumption}
\subsection{Nominal Stabilizing Control}
\label{sec: nom-stab-ctrl}

The nominal learned dynamics $f_{\theta}$ in (\ref{eq:nom-target-traj}) generate the nominal target trajectory $z^*(t)$. However, even under ideal conditions with no task-execution gap, convergence from any initial condition $z_0 \ne z_0^\star$ to $z^{*}(t)$ is not guaranteed unless explicit constraints are imposed on $f_{\theta}$ during training~\citep{figueroa2022locally, sochopoulos2024learning}.
To this end, we augment the nominal learned dynamics with a stabilizing control input $\unom$, forming the regulated learned dynamics:
\begin{equation}
\label{eq:stab-nom}
\dot{z}(t) = f_{\theta}(z(t)) + \unom(z(t), z^*(t)), \quad z(t_0) = z_0.
\end{equation}
The objective of $\unom$ is to ensure that the state $z(t)$ exponentially converges to the target trajectory $z^*(t)$.
For this purpose, 
from (\ref{eq:nom-target-traj}) and (\ref{eq:stab-nom}), 
we analyze the tracking error $e(t) \triangleq z(t) - z^*(t)$, whose dynamics are given by:
$\dot{e}(t) 
      = \left( f_{\theta}(z(t)) + \unom(z(t), z^*(t)) \right) - f_{\theta}(z^*(t)),\ \text{with} \ e(t_0) = z_0 - z_0^\star.$
To render the error dynamics exponentially stable to 0,
we follow Section  \ref{sec:bg-clf}, and select a $C^1$ CLF candidate $V: \mathbb{R}^d \to \mathbb{R}_{\ge 0}$ such that $V(e) > 0$ for $e \neq 0$ and $V(0) = 0$. 
Following \citep{nawaz2024learning}, we obtain the control input $\unom$ as the solution to the following CLF-QP: 
\begin{equation}
\label{eq:unom}
\unom(z(t), z^*(t)) = \arg \min_{u \in \mathbb{R}^d} \quad \frac{1}{2} ||u||^2  \
\text{s.t.} \ \nabla V(e(t))^T \left( f_{\theta}(z(t)) - f_{\theta}(z^*(t)) + u \right) + c V(e(t)) \le 0. 
\end{equation}
This approach enables us to superimpose a stabilizing controller onto any learned $f_{\theta}$.
To characterize the stability of the CLF-based control, we formalize the regularity and boundedness conditions typically satisfied by the CLF-based closed-loop system in the following assumption.
\begin{assumption}[Nominal exponential stability]
\label{ass:clf}
There exist positive constants $\alpha_1, \alpha_2, \lambda$ such that for $e(t) = z(t) - z^*(t)$, with $t \ge t_0$ , $z(t) \in \mathcal{T}(\rho)$: (i) the CLF $V(e)$ is quadratically bounded uniformly in time: $\alpha_1 ||e(t)||^2 \le V(e(t)) \le \alpha_2 ||e(t)||^2$ (ii) the controller $\unom$ ensures an exponential convergence: $\dot{V}(e(t)) \le -2\lambda V(e(t))$ \citep{khalil2002nonlinear}. (iii) $V(e)$ has bounded gradient on the tube: $||\nabla V(e(t))|| \le \Delta_b (\rho)$ (written simply as $\Delta_b$ for brevity hereafter).
\end{assumption}

\subsection{Target Point Selection via Windowed DTW}
\label{subsec:method-dtw}
\begin{figure}[h]
    \begin{minipage}[c]{0.57\columnwidth}
        \captionsetup{width=0.9\linewidth} 
        \begin{algorithm}[H] 
        \noindent \caption{Windowed DTW-Based Target Selector}
        \label{alg:dtw_selector}
        \LinesNumbered
        \SetKwInOut{Input}{Input}
        \SetKwInOut{Output}{Output}
        
        \Input{
            Current state $z_t$; 
            Execution history $Z_{t-H:t}$; \\
            Target trajectory $Z^{*}_{1:N}$;
            Previous index $k_{\mathrm{prev}}$; \\
            Forward window $W$; \\
            Target history $H'$ (default $H'=H$).
        }
        \Output{
            New index $k_{\mathrm{new}}$; 
            Target $z^\star = z^*_{k_{\mathrm{new}}}$
        }
        
        \BlankLine\hrule\BlankLine
        $Z_{\text{hist}} \gets Z_{t-H:t}$ 
        
        \For{$k = k_{\mathrm{prev}}$ \KwTo $\min(N,\,k_{\mathrm{prev}} + W)$}
        {
        $Z^{*(k)} \gets \{z^*_{\max(1, k-H')}, \dots, z^*_k\}$; 
        
        $C[k] \gets \text{DTW}(Z_{\text{hist}}, Z^{*(k)})$}
        $k_{\mathrm{new}} \gets \arg\min_k C[k]$;
         \ $z^\star \gets z^{*}_{k_{\text{new}}}$
        \BlankLine\hrule\BlankLine
        \end{algorithm}
    \end{minipage}
    \hfill 
    \begin{minipage}[c]{0.40\columnwidth}
        \captionsetup{width=0.98\linewidth} 
        \includegraphics[width=0.9\linewidth]{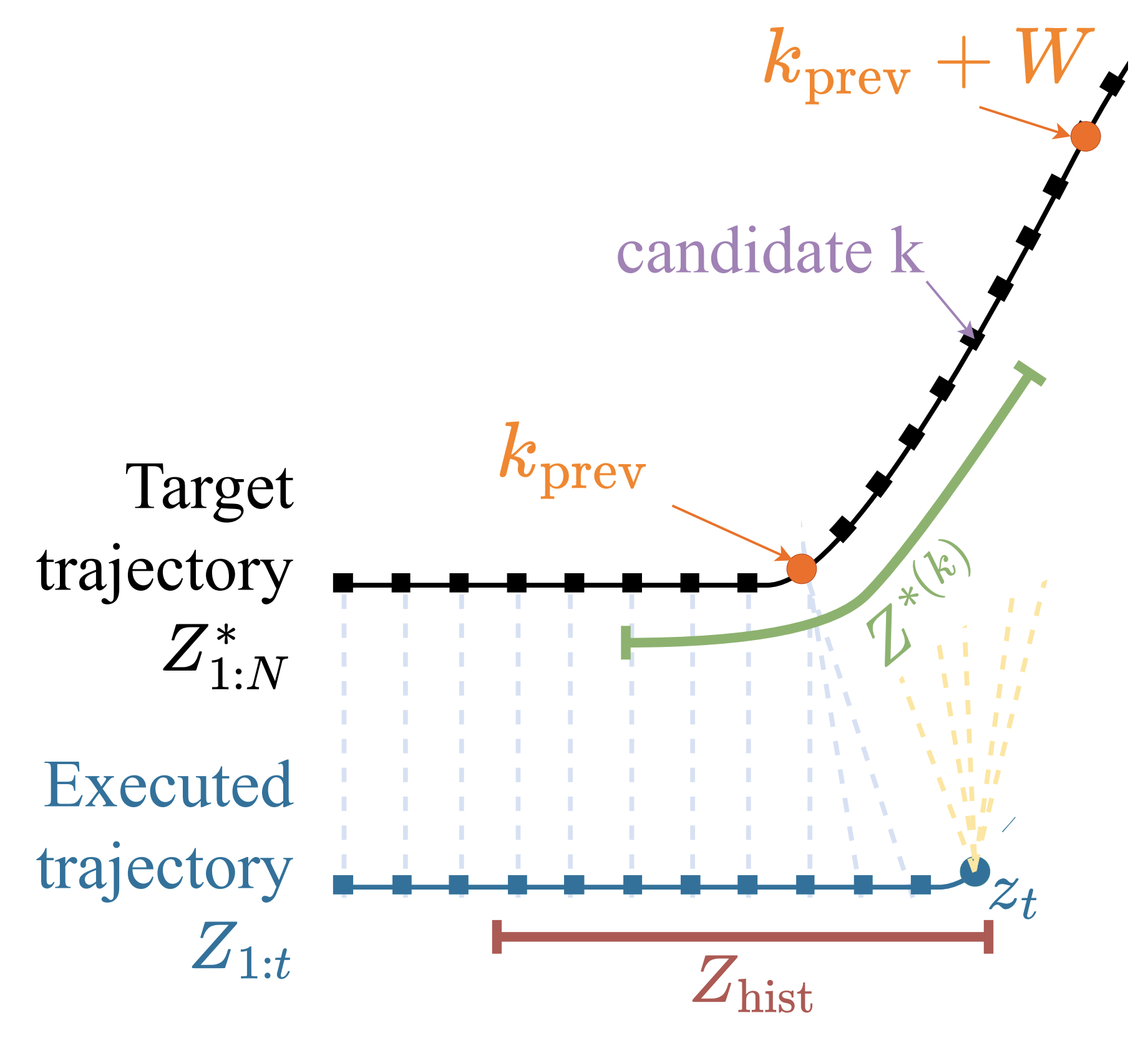}
        \caption{Illustration for how Algorithm \ref{alg:dtw_selector} chooses the target point.}
        \label{fig:algo}
    \end{minipage}
\end{figure}

To converge to the target trajectory $z^*(t)$ in (\ref{eq:nom-target-traj}), the regulated nominal dynamics (\ref{eq:stab-nom}) generates a continuous reference trajectory that needs to be executed by the low-level stack. However, during execution,
the actual
state of the robot $z(t)$ may be phase-shifted relative to the timed target $z^*(t)$ due to disturbances (e.g., suddenly being pushed off-course).
If we use a time-indexed target $z^*(t)$ in (\ref{eq:unom}), and the robot is lagging, the error  $e(t) = z(t) - z^*(t)$ will be artificially large and ``phase-inconsistent." This misalignment might lead to the performance deterioration of the nominal stabilizing controller (\ref{eq:unom}) that takes $z^{*}(t)$ as an input, resulting in  a poor $\unom$ (e.g. trying to ``skip'' parts of the motion to catch up). Therefore, the nominal controller requires a target selection mechanism to provide a phase-consistent target point at each time instant, denoted as $z^{*}$. We propose a windowed DTW-based target selector, detailed in Algorithm \ref{alg:dtw_selector}, and illustrated in Figure \ref{fig:algo}. 

Recall that the nominal learned dynamics, defined by $f_\theta$, generate a continuous target trajectory $z^*(t)$, as given  in (\ref{eq:nom-target-traj}). For the discrete-time implementation of our controller, we sample this trajectory to create a discrete target sequence $Z^{*}_{1:N} \triangleq \{z^{*}_{1},\dots,z^{*}_{N}\}$ of length $N$. At each control step, Algorithm \ref{alg:dtw_selector} aligns the robot's recent execution history, $Z_{\text{hist}} \triangleq Z_{t-H:t} = \{z_{t-H},\dots,z_{t}\}$, with the target trajectory $Z^{*}_{1:N}$. To ensure forward progress, the search for $z^{*}$ is constrained to a forward window of size $W$ (Line 2), starting from the previously selected reference index $k_{\mathrm{prev}} \in \{1,\dots N\}$. For each point $k \in \{k_{\text{prev}},...,k_{\text{prev}}+W\}$ in the forward window (Lines 2-5), the algorithm extracts a candidate target subsequence $Z^{*(k)} \triangleq \{z^*_{\max(1, k-H')}, \dots, z^*_k\}$ (Line 3) and computes the DTW distance, $C[k]$, between $Z_{\text{hist}}$ and $Z^{*(k)}$ (Line 4). After scanning the window, the index $k_{\mathrm{new}}$ that yields the minimum DTW distance is selected (Line 6). This index defines the new target point $z^\star = z^*_{k_{\mathrm{new}}}$, which is then used to compute 
$u_{\text{nom}}(z, z^*)$, and update $k_{\text{prev}}$ for the next control step. This selection mechanism, 
by recomputing the target point based on local geometric similarity, along with the forward-only search, ensures phase-consistent tracking. 

\subsection{\texorpdfstring{$\mathcal{L}_1$}{L1} Adaptive Control Augmentation}

\subsubsection{Task-execution mismatch}
\label{sec:task-exec-mistmatch}
The input \(u_{\mathrm{nom}}\) in (\ref{eq:unom}) is designed to shape the DS flow in the task-space assuming there is no task-execution gap. 
During execution, an inner stack (low-level controller, actuator/plant dynamics, sensing/communication) attempts to realize the regulated nominal motion plan in (\ref{eq:stab-nom}). Imperfections in this realization due to delays, unmodeled dynamics, and model inaccuracies, cause a deviation in the actual dynamics of the task state. These imperfections are modeled at the task-level using a discrepancy term $\sigma(z(t))$, such that the task dynamics is realized as:
\[
\dot{z}(t)
\;=\;
f_\theta(z(t))
\;+\; u_{\mathrm{nom}}(z(t),z^\star(t))
\;+\; {\,\sigma(z(t))}, \quad z(t_0) = z_0.
\]
Due to the nature of the DS-based formulation, the discrepancy term appears as a matched disturbance entering through the same channel as the input $\unom$, motivating the augmentation of $\mathcal{L}_1$ adaptive control. The properties of matched uncertainty with respect to the dynamics are stated in the following assumption.

\begin{assumption}[Bounded uncertainty and derivatives]
\label{ass:unc} 
 The uncertainty $\sigma: \mathbb{R}^d \to \mathbb{R}^d$ is continuous and uniformly bounded by a constant $\Delta_{\sigma} > 0$, and its derivative is uniformly bounded by constant $L_{\sigma z} \ge 0$, i.e.
        $ ||\sigma(z(t))|| \le \Delta_{\sigma} \ \text{and} \ 
        \left|\left|\frac{\partial \sigma}{\partial z}(z(t))\right|\right| \le L_{\sigma z}, 
        \ \text{for all } z(t) \in \mathcal{T}(\rho)$.
    
\end{assumption}

\subsubsection{Closed-loop with \texorpdfstring{$\mathcal{L}_1$}{L1} adaptive control}
\label{section:close_loop_task_level}


We now define the closed-loop task-space system by adding the $\mathcal{L}_1$ adaptive augmentation, $u_a(t) \in \mathbb{R}^d$, to the nominal dynamics and the uncertainty term defined in Section \ref{sec:task-exec-mistmatch}: 
\begin{equation}
\label{eq:closed_loop_l1}
\dot{z}(t) = \underbrace{f_{\theta}(z(t)) + u_{\text{nom}}(z(t), z^*(t))}_{\text{Regulated Nominal Dynamics}} + \underbrace{u_a(t)}_{\mathcal{L}_1 \text{ Augmentation}} + \underbrace{\sigma(z(t))}_{\text{Uncertainty}}, \quad z(t_0) = z_0.
\end{equation}
The adaptive control input $u_a(t)$ is generated by using the $\mathcal{L}_1$ adaptive control architecture described in Section \ref{sec:bg-l1}. Equation~\eqref{eq:closed_loop_l1} is expressed in the same form as the general control-affine nonlinear system in~\eqref{eq:control-affine-nonlinear-bg}, under the following correspondence: the task-space state \( z(t) \) maps to the system state \( x(t) \); the nominal learned dynamics \( f_{\theta}(z(t)) \) map to \( g(x(t)) \); the CLF-based nominal control \( u_{\text{nom}}(z(t), z^*(t)) \) maps to \( u_{\text{nom}}(t) \); and the uncertainty term \( \sigma(z(t)) \) maps to \( \delta(t, x(t), u(t)) \).
With this mapping, the $\mathcal{L}_1$ adaptive control components (state predictor, piecewise-constant adaptive law, and low-pass filter) from Section \ref{sec:bg-l1} can be used to estimate the uncertainty $\hat{\sigma}(t) \in \mathbb{R}^d$ and generate the input $u_a(t)$. 
Hence, following (\ref{eq:closed_loop_l1}), at every time instant, our $\mathcal{L}_1$-DS-based architecture computes the reference point $z_{\text{ref}}$, for the low-level stack to track, by solving, 
$\dot{z}_{ref}(t) = f_\theta(z(t)) + u_{\text{nom}}(z(t),z^*(t)) + u_a(t) \ \text{with} \ z_{\text{ref}}(t_0) = z_0. $

Following \citet{wu2023mathcal}, using Assumptions~\ref{ass:domain}--\ref{ass:unc},
by continuity on the compact set $\mathcal T(\rho)$, there exist finite constants $\Delta_f,\Delta_{\mathrm{nom}}$ such that:
$\|f_\theta(z(t))\|\le \Delta_f,
\|u_{\mathrm{nom}}(z(t),z^\star(t))\|\le \Delta_{\mathrm{nom}} \ \forall z(t)\in\mathcal T(\rho).$ 
 Note, the bound $\Delta_{nom}$ is a consequence of the continuity of the mapping $(z, z^*) \mapsto u_{nom}$ over the compact set $\mathcal{T}(\rho)$ rather than an explicit optimization constraint. We also define $\Delta_{\hat\sigma}$ as the uniform bound on the uncertainty estimate:
$\Delta_{\hat\sigma}:=\sup_{t\ge t_0}\|\hat\sigma(t)\|$.
Hence, the bounds on the state derivative in (\ref{eq:closed_loop_l1}) are given by
$\|\dot z\|_\infty \le \phi_1 \triangleq \Delta_f + \Delta_{\mathrm{nom}} + \Delta_{\hat\sigma} + \Delta_\sigma.$
The Huruwitz matrix in (\ref{eq:state_predictor_bg}) is chosen diagonal, $A_s = \mathrm{diag(\lambda_1,\hdots,\lambda_k)}, \ \text{where} \ \lambda_k \ \text{for} \ k \in \{1, \dots d\}$ are negative scalars. 
Based on the above bounds and constants from Assumptions~\ref{ass:domain}--\ref{ass:unc}, we define the following constants :
$\zeta_1(\omega)\triangleq\frac{\Delta_\sigma}{|2\lambda - \omega|} + \frac{L_{\sigma z}\,\phi_{1}}{2\lambda\,\omega},$
$\zeta_2(A_s) \triangleq 2\sqrt{d}\,L_{\sigma z}\,\phi_1\ +\ \sqrt{d}\,\max_k|\lambda_k(A_s)|\,\Delta_\sigma,$
$\zeta_3(\omega) \triangleq\Delta_\sigma\,\omega,$
and $\zeta_4(\omega, A_s)  \triangleq \frac{\Delta_b (\zeta_2(A_s) + \zeta_3(\omega))}{2\lambda}. $
For an arbitrarily chosen scalar $\varepsilon>0$, we define the tube radius in (\ref{eq:goal}) as
$\rho \;\triangleq\; \|e(t_0)\|\,\sqrt{\frac{\alpha_2}{\alpha_1}}\;+\;\varepsilon,$
which implies $V(e(t_0)) \triangleq V_0 < \alpha_1 \rho^2$. 
The $\mathcal{L}_1$ filter bandwidth $\omega > 0$ is chosen large enough, and the sampling time $T_s > 0$ is chosen small enough, satisfying the following conditions:
\begin{equation}
\alpha_1\,\rho^2 > V_0 + \Delta_b\,\zeta_1(\omega), \quad
T_s \le \frac{\alpha_1 \rho^2 - V_0 - \Delta_b \zeta_1(\omega)}{\zeta_4(\omega, A_s)}.
\label{eq:design}
\end{equation}
Conditions (\ref{eq:design}) can be met for large $\omega$ since $\zeta_1(\omega) = O(\omega^{-1})$ and $V_0 < \alpha_1 \rho^2$, which
guarantees a valid 
$T_s > 0$. With the above constants and design conditions in place, we can now formally state the main result guaranteeing that $z(t)$ remains bounded around $z^{*}(t)$.

\begin{theorem}
\label{thm:main}
Consider the dynamical systems in (\ref{eq:nom-target-traj}) and (\ref{eq:closed_loop_l1}). Let Assumptions~\ref{ass:domain}--\ref{ass:unc} and inequalities \eqref{eq:design} hold. Then $z(t)$ is bounded around $z^{*}(t)$, $z(t) \in \mathcal{O}(z^*(t), \rho), \ \text{for all}\ t \ge t_0$. Furthermore, the closed-loop state $z(t)$ in~\eqref{eq:closed_loop_l1} is uniformly ultimately bounded $\text{for all}\ t \geq t_1 > t_0$, such that 
\[
z(t) \in \mathcal{O}(z^*(t), \mu(\omega,T_{s},t_{1})) \subset \mathcal{O}(z^*(t), \rho), 
\]
where the ultimate bound is defined as 
\[
\mu(\omega,T_{s},t_{1}) := \sqrt{\frac{e^{-2\lambda (t_1-t_{0})}V_0\ +\ \Delta_b \zeta_1(\omega) + \zeta_4(\omega, A_s) T_s}{\alpha_1}}.
\]
\end{theorem}

\noindent The result follows from Theorem 1 in \citet{wu2023mathcal} and establishes that the proposed $\mathcal{L}_1$-DS architecture guarantees uniform bounds on the error between $z(t)$ and $z^*(t)$.

\section{Simulations}

Having established the $\mathcal{L}_{1}$-DS architecture, we now empirically validate its performance
using a neural ordinary differential equations (NODE) model trained on the non-periodic LASA, and periodic IROS handwriting datasets \citep{sochopoulos2024learning, nawaz2024learning}. We demonstrate the efficacy of $\mathcal{L}_1$-DS in two execution regimes. (i) \emph{Perfect Command Following}, which simulates a perfect low-level executor, 
i.e., the input to the forward kinematics block is the same as the output of the inverse kinematics block in Figure \ref{fig:architecture}. In this regime, disturbances are introduced directly into task dynamics via $\sigma$ in (\ref{eq:closed_loop_l1}) to demonstrate the robustness of the task-level planning layer. (ii) \emph{Imperfect Command Following}, which models the task-execution mismatch where our low-level plant and controller cannot perfectly realize task-level commands. This regime considers second-order closed-loop plant dynamics at the low-level and is subject to matched (M) and/or unmatched (U) disturbances at the low-level (see Appendix \ref{appendix:expt_continued}).

 In Table \ref{tab:node-full}, performance is reported using the DTW distance between the executed task-space trajectory $z(t)$ and the nominal target trajectory $z^*(t)$. DTW scores are sensitive to the specific task's (shape's) length and complexity, as each DS reacts differently to the same disturbance type. Therefore, instead of reporting the mean and variance of the DTW across the entire datasets, we normalize the scores to better assess relative improvements. For each disturbance type and shape, we divide the DTW score obtained by the CLF augmented, NODE+CLF \citep{nawaz2024learning}, and $\mathcal{L}_1$-DS ($\mathcal{L}_1$-NODE) versions, by the score obtained from its learned nominal model (NODE).
 This normalization yields scores typically around or below 1.0, where values less than 1.0 indicate a performance improvement attributable to the control augmentation.

\begin{figure}[t]
    \centering
    \includegraphics[width=\columnwidth]{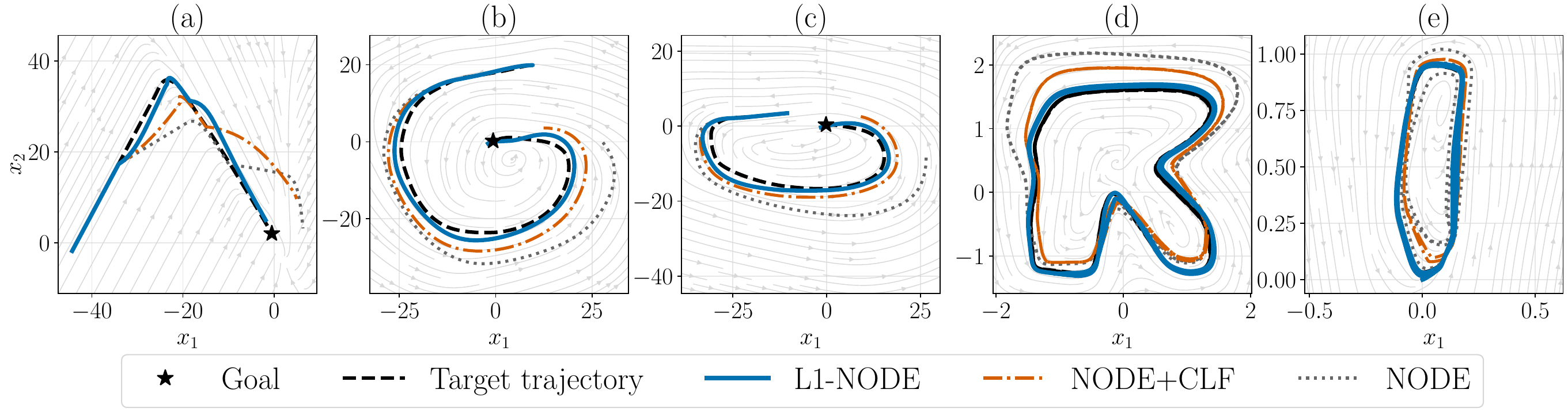}
    \caption{(a) LASA \textit{Angle}, perfect command following regime with step disturbance; (b) LASA \textit{GShape}, imperfect command following regime with unmatched multi-sine disturbance; (c) LASA \textit{DoubleBendedLine}, imperfect command following regime with matched multi-sine disturbance; (d) IROS \textit{RShape}, imperfect command following regime with unmatched constant disturbance; (e) IROS \textit{IShape}, imperfect command following regime with matched multi-sine and unmatched periodic step disturbance.}
    \label{fig:lasa}
\end{figure}



\begin{table*}[htbp]
\centering
{\mytextsize 
\caption{Average Normalized DTW Scores. Lower score is better (1.0 = NODE performance).}
\label{tab:node-full}
\begin{tabular}{@{} l c c c c c c @{}}
\toprule 
& \multicolumn{2}{c}{Dataset: LASA} & \multicolumn{3}{c}{Dataset: IROS} \\
 \cmidrule(lr){2-3} \cmidrule(lr){4-6}
Regime: Disturbance & NODE+CLF & \textbf{$\mathcal{L}_1$-NODE} & NODE+CLF(LE) & NODE+CLF(DTW) & \textbf{$\mathcal{L}_1$-NODE} \\
\midrule
Perfect: Step    & 0.627 $\pm$ 0.026 & \textbf{0.118 $\pm$ 0.002} & 0.717 $\pm$ 0.043 & 0.693 $\pm$ 0.067 & \textbf{0.396 $\pm$ 0.022} \\
--- Imperfect: ---   &  &  &   &  &  \\
M. Multi-sine & 0.707 $\pm$ 0.118 & \textbf{0.244 $\pm$ 0.023} & 0.720 $\pm$ 0.078 & 0.648 $\pm$ 0.147 & \textbf{0.536 $\pm$ 0.116} \\
U. Constant   & 0.628 $\pm$ 0.031 & \textbf{0.196 $\pm$ 0.010} & 0.435 $\pm$ 0.011 & 0.394 $\pm$ 0.011 & \textbf{0.042 $\pm$ 0.003} \\
U. Multi-sine& 0.694 $\pm$ 0.103 & \textbf{0.237 $\pm$ 0.019} & 0.585 $\pm$ 0.013 & 0.482 $\pm$ 0.020 & \textbf{0.327 $\pm$ 0.020} \\
M. Multi-sine with  &  &   &  &  &  \\
U. Pulses& 0.687 $\pm$ 0.065 & \textbf{0.241 $\pm$ 0.019} & 0.630 $\pm$ 0.003 & 0.482 $\pm$ 0.020 & \textbf{0.212 $\pm$ 0.016} \\
\bottomrule
\end{tabular}
} 
\end{table*}






On the IROS dataset, NODE+CLF(LE) uses the least-effort-based target selector \citep{nawaz2024learning} designed for periodic tasks, while NODE+CLF(DTW) uses our proposed Algorithm \ref{alg:dtw_selector} for selecting the target point. We see that the proposed Algorithm \ref{alg:dtw_selector} improves the performance of the nominal controller (\ref{eq:unom}), which is further improved by the augmentation of $\mathcal{L}_1$ adaptive control. Overall, for both datasets, the $\mathcal{L}_1$-DS architecture improves the tracking performance under the perfect and imperfect execution regimes in the presence of disturbances. In Figure~\ref{fig:lasa}, we show sample experiments corresponding to Table~\ref{tab:node-full}. Further simulations with a SEDS \citep{khansari2011learning} nominal model can be found in  Appendix \ref{appendix:seds}. 



\section{Conclusion}
We introduce $\mathcal{L}_{1}$-DS, a task-level control architecture designed to address the task-execution mismatch in DS-based motion plans.
The proposed architecture leverages the dynamical system nature of the DS-based motion plan formulation to augment control-theoretic methods for stability and robustness. 
We also design a novel windowed DTW-based target selector for shape-consistent tracking.
We successfully demonstrate the efficacy of the proposed architecture on periodic and non-periodic handwriting tasks. 
Future work includes experimental validations on higher dimensional task spaces and robotic manipulators, robustness to errors in the learned nominal DS-based model, and extensions to multiple skills or skills involving sequential sub-tasks. 


\section{Acknowledgments}
This work is supported by the Air Force Office of Scientific Research Grant FA9550-21-1-0411, the National Aeronautics and Space Administration under Grant 80NSSC22M0070, the National Science Foundation (NSF) under Grants CMMI 2135925, IIS 2331878, and  CMMI 24-31216, and the Swiss National Science Foundation under grant agreement P500PT$\_$230223.


\bibliography{references}

\newpage

\appendix

\section{Simulations with a SEDS model}
\label{appendix:seds}

\begin{table*}[htbp]
\centering
{\mytextsize 
\caption{Average Normalized DTW($z$, $z^*$). Lower score is better (1.0 = SEDS performance).}
\label{tab:node-comparison}
\begin{tabular}{@{} l c c @{}}
\toprule 
& \multicolumn{2}{c}{Dataset: LASA} \\
 \cmidrule(lr){2-3} 
Regime: Disturbance & SEDS+CLF & \textbf{$\mathcal{L}_1$-SEDS}  \\
\midrule
Perfect: Pulses    & 0.080 $\pm$ 0.000 & \textbf{0.014 $\pm$ 0.000}  \\
--- Imperfect: ---  &  &  \\
M. Multi-sine & 0.256 $\pm$ 0.012 & \textbf{0.011 $\pm$ 0.003} \\
U. Constant   &  0.322 $\pm$ 0.051& \textbf{0.028 $\pm$ 0.000} \\
U. Multi-sine& 0.447 $\pm$ 0.026 & \textbf{0.258 $\pm$ 0.008} \\
M. Multi-sine with &  &  \\
U. Pulses& 0.264 $\pm$ 0.015 & \textbf{0.067 $\pm$ 0.001} \\
\bottomrule
\end{tabular}
} 
\end{table*}

We also validate the efficacy of $\mathcal{L}_1$-DS using a SEDS \citep{khansari2011learning} nominal model. SEDS is designed to generate inherently stable DS-based motion plans for convergence to a single target attractor point in the task-space. As discussed in Section \ref{subsec:method-dtw}, this does not guarantee convergence to a nominal target trajectory.  Therefore, we augment it with a CLF-based controller for a fair comparison for the objective of shape trajectory-tracking.

\section{Simulation Setup}
\label{appendix:expt_continued}

\subsection{Datasets and Preprocessing}

\paragraph{LASA (non-periodic) shapes.}
We use 10 randomly chosen LASA shapes (Angle, CShape, DoubleBendedLine, GShape, PShape, Sshape, Sine, WShape, Worm, Snake). 
All demonstrations are resampled to a common, normalized time grid
$t \in [0,1]$ with $N=1000$ samples, unless stated otherwise. We use 4 demonstrations for training a NODE for each shape. The initial state for all runs is
$[\bar z(t_0)]$, the average starting point of the 4 training demonstrations. 

\paragraph{IROS (periodic) shapes.}
For R/I/O/S shapes, demonstrations are resampled to $N=10{,}000$ points on $t\in[0,1]$.
We use $n_{\text{train}}=3$ demos for training (if available).
The outer simulation step is $\Delta t = 1/(N-1)$.
The initial state is $[\bar z(t_0)]$.

\subsection{Learned Task Models}

\paragraph{Neural ODE (NODE).}

\emph{IROS (NODE).}
Shapes: RShape, IShape, OShape, SShape. Training: $50$k steps; batch $16$; base learning rate $5\times10^{-4}$. MLP architecture: width $128$, depth $3$.
Resampling: $N=10{,}000$ points.
\emph{LASA (NODE).}
MLP architecture: width $128$, depth $3$.
Resampling: $N=1000$ points.



\paragraph{SEDS (LASA ablations).}
A stable DS (SEDS) is fitted with $K$ Gaussians:
$K=8$ for \textit{Sshape}, \textit{Sine}, \textit{Snake}, and $K=6$ for the other LASA shapes.
Training and evaluation use $N=1000$ resampled points.

\subsection{Target Selection Policies}

Two target selection policies are evaluated:
\begin{enumerate}
  \item \textbf{DTW-based selector.} Forward-only alignment to the NODE reference with local windows
  $W=50$ and $H=40$; initialized at $\bar z(t_0)$.
  \item \textbf{Least-Effort selector.} Look-ahead over a horizon of $N_{\mathrm{LA}}=35$
  steps on $\Delta t$. The starting state is $\bar z(t_0)$.
\end{enumerate}

\subsection{Controller Stack and Simulation Modes}

We compare three task-level controllers: \emph{NODE/SEDS:} tracking using only the learned field $f_\theta$. \emph{NODE/SEDS+CLF:} NODE/SEDS augmented with a CLF-QP nominal stabilizer. \emph{L1-NODE/SEDS:} NODE/SEDS+CLF additionally augmented with an $\mathcal{L}_1$ adaptive control input (we use windowed-DTW for target point selection, unless specified).


\paragraph{Two simulation regimes}
\begin{enumerate}
  \item \textbf{Perfect command tracking (direct task-space integration).}
  $\sigma$ is injected directly in task space. We use two rectangular pulses on the
  normalized time axis. In Figure \ref{fig:lasa_dist}(a), $\sigma = [\sigma(x_1), \sigma(x_2)]^\top$ are the disturbances along $z(t)=[x_1,x_2]^\top$ axes.
  This experiment isolates task-level robustness.
  \item \textbf{Imperfect command tracking (plant + low-level controller).}
    We model the lower-level \emph{plant} in task space with position $p(t)=[p_1,p_2]^\top$ and velocity $v(t)=[v_1,v_2]^\top$. Disturbances are split into
    matched $d_m(t)=[d_{m,1},d_{m,2}]^\top$ (entering the control channel) and unmatched $d_{um}(t)=[d_{um,1},d_{um,2}]^\top$ (entering the position–rate channel directly). Per axis $i\in\{1,2\}$ the double–integrator plant is
    \begin{align}
    \dot p_i(t) &= v_i(t)\;+\;d_{um,i}(t), \label{eq:plant-pos-um}\\
    \dot v_i(t) &= u_i(t)\;+\;d_{m,i}(t), \label{eq:plant-vel-m}
    \end{align}
    or, in vector form,
    \[
    \dot p(t) = v(t) + d_{um}(t), \qquad
    \dot v(t) = u(t) + d_m(t),
    \]
    with control input $u(t)=[u_1,u_2]^\top$ produced by a low–level PID acting independently on each axis to track a reference $(p_r(t),v_r(t))$ supplied by the task–level selector:
    \begin{align}
    e_i(t)&=p_{r,i}(t)-p_i(t), \qquad \dot e_i(t)=v_{r,i}(t)-v_i(t),\\
    u_i(t)&=K_p\,e_i(t)+K_d\,\dot e_i(t)+K_i\!\int_0^t e_i(\tau)\,d\tau .
    \end{align}
    The selector (NODE/CLF/$\mathcal L_1$ stack) updates $(p_r,v_r)$ at the \emph{outer} simulation step $\Delta t$, while the plant+PID loop runs $10\times$ faster with inner step $\Delta t_{\text{llc}}=\Delta t/10$. Thus, task-level architecture only shapes $(p_r,v_r)$, and the PID enforces fast, stable tracking on the double–integrator in the presence of unmatched $d_{um}$ and matched $d_m$ disturbances.

\end{enumerate}


\subsection{Disturbances}

\begin{figure}[t]
    \centering
    \includegraphics[width=\columnwidth]{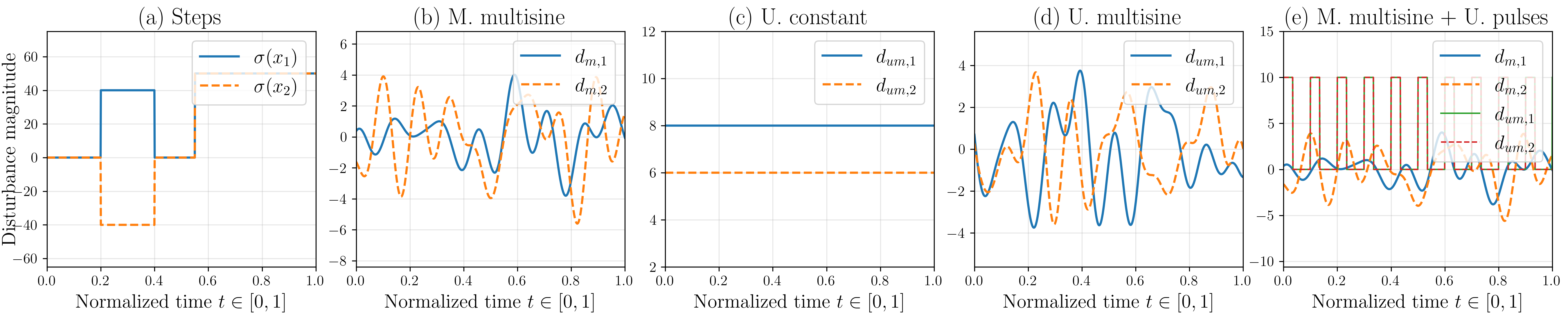}
    \caption{Disturbances used for the LASA dataset experiments. Appropriate magnitude-scaled versions of these were used for the IROS experiments.}
    \label{fig:lasa_dist}
\end{figure}

All disturbance signals are plotted on the normalized time grid as shown in Figure \ref{fig:lasa_dist}.

\end{document}